\documentclass[runningheads]{llncs}
\usepackage[utf8]{inputenc}
\usepackage{authblk}
\usepackage{graphicx}
\usepackage{microtype}
\usepackage{booktabs}
\usepackage{makecell}
\usepackage{amsmath}
\usepackage[section]{placeins}
\usepackage{setspace}
\usepackage{titletoc}
\usepackage{marvosym}

\begin{document}  

\begin{sloppypar}%防止行溢出页边距
\begin{spacing}{1}

\title{Federated Self-supervised Domain Generalization for Label-efficient Polyp Segmentation}

\def\MICCAISubNumber{26}
\titlerunning{ADSMI-24 submission ID \MICCAISubNumber} 
\authorrunning{ADSMI-24 submission ID \MICCAISubNumber} 

\author{Xinyi Tan}
\author{Jiacheng Wang}
\author{Liansheng Wang \textsuperscript{\Letter}}
\affil{Department of Computer Science at School of Informatics, Xiamen University, Xiamen 361005, China \authorcr
          \{xinyitan, jiachengw\}@stu.xmu.edu.cn, lswang@xmu.edu.cn }

\institute{}

\maketitle              % typeset the header of the contribution
\begin{abstract}
% Automatic detection of intestinal polyps during colonoscopy holds profound implications for cancer prevention, treatment efficacy, and patient prognosis. 
Employing self-supervised learning (SSL) methodologies assumes par-amount significance in handling unlabeled polyp datasets when building deep learning-based automatic polyp segmentation models. However, the intricate privacy dynamics surrounding medical data often preclude seamless data sharing among disparate medical centers. Federated learning (FL) emerges as a formidable solution to this privacy conundrum, yet within the realm of FL, optimizing model generalization stands as a pressing imperative. Robust generalization capabilities are imperative to ensure the model's efficacy across diverse geographical domains post-training on localized client datasets.
In this paper, a \textbf{F}ederated self-supervised \textbf{D}omain \textbf{G}eneralization method is proposed to enhance the generalization capacity of federated and \textbf{L}abel-efficient intestinal polyp segmentation, named LFDG. 
Based on a classical SSL method, DropPos, LFDG proposes an adversarial learning-based data augmentation method (SSADA) to enhance the data diversity. 
LFDG further proposes a relaxation module based on Source-reconstruction and Augmentation-masking (SRAM) to maintain stability in feature learning.
We have validated LFDG on polyp images from six medical centers. The performance of our method achieves 3.80\% and 3.92\% better than the baseline and other recent FL methods and SSL methods, respectively.

% The incorporation of a data augmentation technique rooted in adversarial learning yielded a significant augmentation in the segmentation accuracy.

\keywords{Federated Learning \and Self-supervised Domain Generalization}
\end{abstract}
\section{Introduction}
Accurate polyp segmentation is crucial for medical diagnosis and prevention, but the scarcity of data labels is a major challenge. Various self-supervised training approaches, including image-only pretext tasks\cite{doersch2015unsupervised,pathak2016context,zhang2016colorful} and contrastive learning\cite{chen2020simple,he2020momentum}, have been proposed to address this issue. Recent research\cite{qu2022rethinking} indicates that pre-training with Vision Transformer in federated learning surpasses traditional convolutional neural network methods. Additionally, pre-training methods like MAE \cite{he2022masked}, BEiT \cite{bao2021beit}, DropPos \cite{wang2024droppos} and other image reconstruction techniques based on Vision Transformer notably diminish the reliance on labeled data.

Federated learning is a promising approach for collaborative model training across distributed data sources, addressing concerns about data privacy and ownership while improving generalization capabilities. Given the privacy challenges in polyp data segmentation, applying federated learning to this area is especially relevant. Several scholarly works~\cite{pfitzner2021federated,rehman2023federated,dayan2021federated} are currently exploring the feasibility and effectiveness of using federated learning in the medical domain.

However, federated learning, as a distributed learning method, typically experiences a decrease in performance compared to the training accuracy of a single-source domain dataset due to data heterogeneity and privacy protection limitations. Regarding polyp data, the variance in data collection protocols and target populations across medical centers results in non-identical distributions and heterogeneity within each center's dataset. Consequently, enhancing the domain generalization capability of federated models has emerged as a pressing concern. Several studies have proposed solutions to domain generalization, such as leveraging semantic spaces \cite{dou2019domain} and employing Meta Learning techniques \cite{li2018learning}. Nevertheless, FL is inherently constrained by privacy preservation measures, precluding the transfer of private information between disparate domains. Drawing inspiration from single-source domain generalization approaches \cite{chen2023meta,krull2019noise2void}, enhancing the generalization capacity at each server-side can significantly enhance the client model's performance on target datasets. Among these methodologies, data augmentation via adversarial learning \cite{volpi2018generalizing,qiao2020learning,zhao2020maximum,chen2022enhancing} serves to expand the training dataset by generating synthetic samples with diverse domain characteristics. 

In this paper, we propose a Federated self-supervised Domain Generalization method to enhance the generalization capacity of the federated learned representations with self-supervision, named LFDG. Basically, it employs the SSL method on each distributed site and aggregates the learned parameters on the server using average updates. Besides, LFDG proposes a self-supervised adversarial data augmentation method based on maximizing perturbation loss and minimizing DropPos pre-training loss to generate enhanced samples and maintain stable representations. To further refine model generalization capabilities, LFDG proposes a Source-reconstruction and Augmentation-masking (SRAM) relaxation module to constrain the perturbation loss to prevent distortion.
We have validated LFDG on an unknown domain by comparing the performance of the proposed method with previously existing joint self-supervised training, and our method have achieved the highest Mean IoU of 62.83\%. Ablation experiments are performed at the same time. The SSADA method and SRAM module can improve the model performance by 2.50\% and 1.30\% respectively.

\section{Methods}
\subsection{Overall Framework}
Our objective is to develop a federated learning framework and enhance its efficacy with medical datasets. Assuming a scenario with N clients, indexed as $k\sim{1,..., N}$, each possessing a local dataset denoted as $D_{k}$ governed by distinct distributions $P_{k}$. Individual clients undergo training on their respective local datasets, leading to the development of a global model denoted as $D_{G}$, achieved through parameter aggregation methodologies. Notably, this process employs federated learning algorithms, such as FedAvg.

\begin{figure}[t]
\centering
\includegraphics[width=\textwidth]{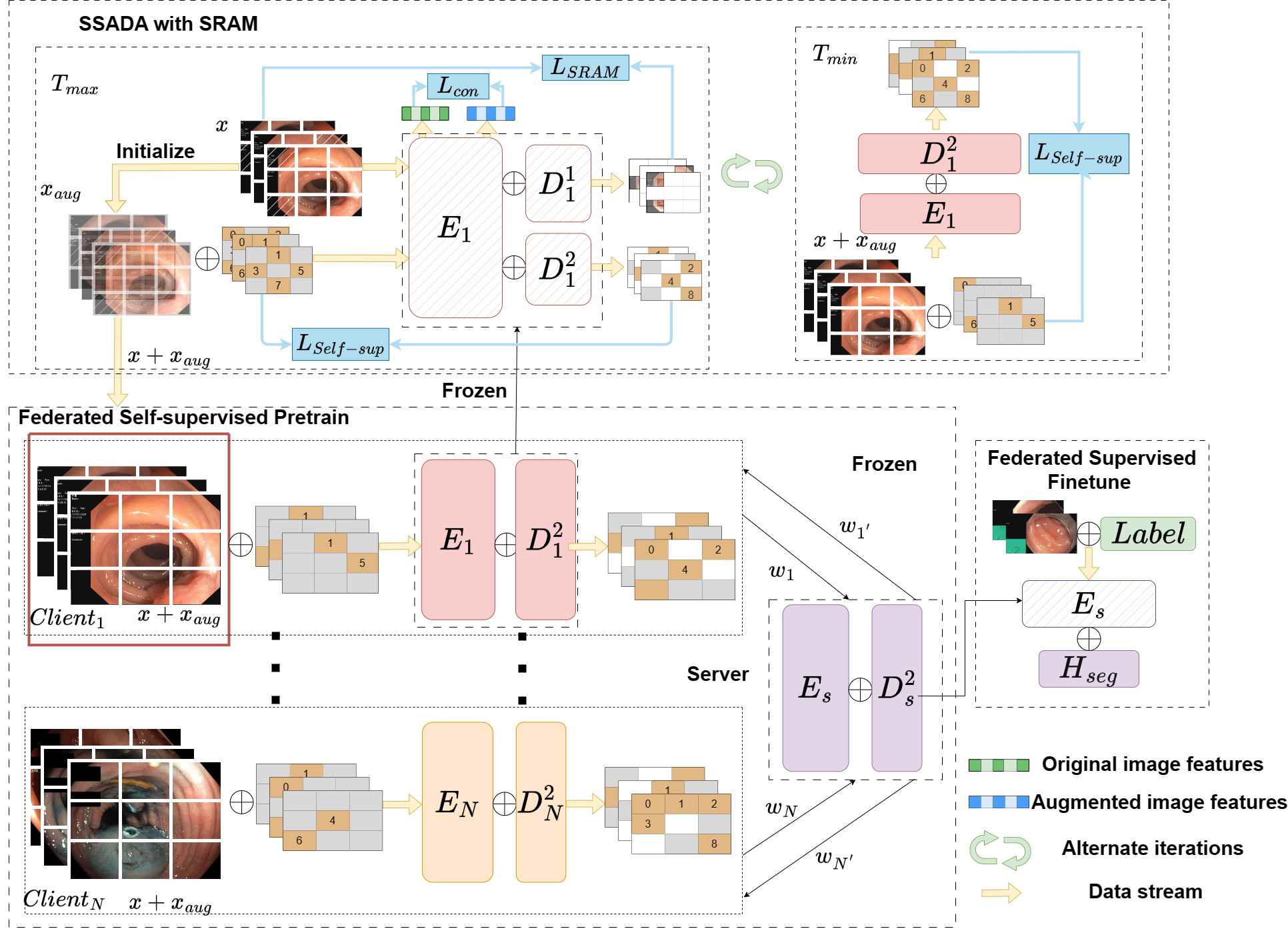}
\caption{The overall framework of LFDG. It learns generalizable representations through self-supervised adversarial data augmentation (SSADA) and source-reconstruction and Augmentation-masking(SRAM) during the local training step and updates the learned parameters in the server.} \label{fig1}
\end{figure}

The LFDG aims to enhance the generalization ability of the federated self-supervised learning paradigm. The basic scenario involves adopting a standard SSL approach in each round of local training, where the learned representation relies on the local data of each site, it utilizes a basic SSL approach such as DropPos, while combining self-supervised adversarial data augmentation and a specialized relaxation module to enhance data diversity while maintaining stable expression, and then uses Fedavg to aggregate model parameters, get a server-side pre-trained model, and finally freeze the pre-trained backbone parameters on the server side, Use local data to fine-tune downstream tasks. The overall framework of LFDG is shown in Figure ~\ref{fig1}.

% After each round of pre-training of the client model, the FedAvg federated learning method is utilized to aggregate parameters, yielding the server-side model. Subsequent to FedAvg pre-training, the server immobilizes the parameters of the Vision Transformer (VIT) and leverages a limited quantity of locally labeled data for fine-tuning of downstream segmentation tasks. Subsequently, the model is evaluated on the validation set.

\subsection{Self-supervised Adversarial Data Augmentation }
% Privacy concerns in federated learning and the diverse nature of multi-center polyp data emphasize the need to improve the generalization capabilities of pre-trained models. This study aims to enhance image representations through adversarial learning, creating a model resilient to perturbations and with stronger generalization. 
The adversarial data augmentation involves two essential phases: maximization and minimization. In the maximization phase, the goal is to increase the discrepancy between the augmented image and its original version, inducing significant perturbations that signify notable domain shifts. In the subsequent minimization phase, iterative model refinement takes place by incorporating augmented images into the dataset and stochastically updating model parameters. This iterative process of maximizing and minimizing leads to the development of a model with enhanced generalization capabilities.
\subsubsection{Maximization Phase}
In order to learn generalizable representations, it is necessary to ensure that the model can have good robustness (that is, good accuracy) for any data in the worst case. Hence, we define the distance between the augmented distribution $P_{aug}$ and the source distribution $P$ as :
\begin{equation}
D_\theta(P_{aug},P) := \mathop{inf}\limits_{M \in {\Pi (P_{aug},P)}} {E_M[c_\theta(X,X_{aug}) ]},
\end{equation}where $c_{\theta}$ represents the distance cost from $X_{aug}$ to $X$, and \begin{equation}
c_\theta(X_{aug},X):= \|z_{aug} - z \|_2^2.
\end{equation}
where $z_i$ and ${z_{aug}}$ denote the feature representations of the original image and the augmented image, respectively. 
The major aim is to enlarge the distance and reduce the original pretraining loss in the meanwhile, gaining useful and generalizable representations. Therefore, the objective can be defined as:
\begin{equation}
    \underset{\theta \sim \Theta}{\text{minimize}} \Big\{F(\theta):= \underset{P}{\text{sup}}\big\{E[L(\theta;X)]-\gamma D_\theta(P_{aug},P)\big\}\Big\},
\end{equation}
where $L(\theta;)$ denotes the self-supervised pertaining loss, i.e., DropPos loss.
Importantly, unlike previous adversarial data augmentation methods, our approach eschews reliance on labeled data for pre-training, the symbol $L_{Self-sup}$ in the adjusted formula indicative of the loss incurred during self-supervised pre-training:

\begin{equation}
\begin{aligned}
&L(\theta;)=L_{Self-sup}=\\
&-\sum\limits_{i=0}^{(1-\gamma)N-1}\sum\limits_{j=0}^{N-1}(1-M_{pos}^i) \cdot  one\_{hot}({pos}_{ij})\cdot log[\frac{exp(f_{ij})}{\sum_{k=0}^{(1-\gamma)N-1}exp(f_{ik})}],
\end{aligned}
\end{equation}
where $f_{ij}$  is the position prediction of the i-th patch, ${pos}_{ij}$ is the real position information, $N$ is the total number of patches, $M_{pos}^i$ is the position mask of 0-1, 1 means that the patch has position embedding and should not be reconstructed, and 0 means that the patch does not have position embedding and needs to be reconstructed.
Given our optimization objective to preserve the original input image $X$, an increase in the reconstruction loss $L_{Self-sup}$ during the maximization process gradually alters the image from its initial state. This alteration helps reduce image heterogeneity, ultimately enhancing overall image quality.

In addition, to address the potential distortion in generated images and enhance generalization performance, we incorporate semantic consistency using Lagrangian relaxation. This approach employs a basic Mean Square Error (MSE) loss function to maintain semantic fidelity: 
\begin{equation}
    c_\theta=L_{con} = \frac{1}{n}\sum\limits_{i=1}^{n}(z_{aug}-{z_i})^2,
\end{equation}
where $n$ represents the number of pairs consisting of original images and their corresponding augmented images. $L_{con}$ diminishes potential distortions by constraining the heterogeneity between the original image and its augmented counterpart.
\subsubsection{Minimization Phase}
All augmented images generated by the maximization stage are merged into the dataset, and the model is iterated for $T_{min}$ rounds to update the parameters. The main idea here is to iteratively learn enhanced data features from a fictitious target distribution while preserving the semantic features of the original data points.

\subsection{Source-reconstruction and Augmentation-masking}
In order to mitigate potential performance degradation resulting from significant noise and excessive distortion within the image, we introduce an additional relaxation module SRAM (Source-reconstruction and Augmentation-masking).

Considering that existing adversarial data augmentation has effectively utilized the patches that were masked in the position-mask part during the pre-training, however the patches that were masked in the image-mask part have not yet been used. Drawing inspiration from the classical Masked Autoencoders methodology, we integrate random mask image reconstruction to quantify the dissimilarity between the original and augmented images, thereby reducing this difference throughout the training stage. Initially, the augmented image $x_{aug}$ undergoes segmentation into a sequence of image patches:$p_{aug}=\{p_{i}\}^n_{i=1}\sim R^{n\times (S^2\cdot C)}$.
$p_{aug}$ is divided into $p_{aug-mask}$ and $p_{aug-unmask}$. Here, $C$ denotes the number of image channels, while $H$ and $W$ represent the height and width of the image, respectively. $S$ signifies the patch size, and $n=W*H/S^2$ corresponds to the total number of patches. The patch is randomly masked with a certain probability, that is, image-mask part. The unmasked patches $p_{aug-unmask}$ are fed into the encoder of pre-training part to obtain the encoded features, and the decoder then reconstructs the entire image to yield the restored pattern $X_{rec}$, which contains $p_{rec}$. Since $p_{rec}$ is reconstructed for the augmented image patches $p_{aug-mask}$, and our goal is to limit the deviation between $X_{rec}$ and the original image $X$. Thus, we compute the loss of the corresponding patches $p_{mask}$ in $X$ against its counterpart $p_{rec}$ in the reconstructed image $X_{rec}$. The SRAM loss, employing a straightforward MSE loss function, is continuously optimized to maximize disturbance during the constraint generation phase:
$L_{SRAM}(\theta;X,X_{aug}) = \frac{1}{n}\sum\limits_{i=1}^{n}({p_{rec}^i}-{p_{mask}^i})^2$.
here, $p_{rec}^i=f(X_{aug})$ denotes the patches reconstructed after encoder and SRAM decoder $D_{SRAM}$ of pre-training, while $p_{mask}^i$ represents the patches of the original image.

Through iterative optimization of the SRAM loss function, the objective of maximizing perturbation during the constraint generation stage is attained.

\subsection{Loss Function}

Our loss function is divided into different stages: perturbation maximization loss and model minimization loss. Specifically, in the k-th stage:
In the perturbation loss maximization stage, we aim to maximize the perturbation loss while taking into account semantic consistency and adhering to the constraints imposed by the SRAM relaxation module. We calculate the adversarial perturbation $X_\gamma ^*$ of the original image $X$ with respect to the current model $\theta$:
\begin{equation}X_\gamma ^* = {argmax}_{X\in \chi }\{L_{Self-sup}(\theta;X) - c_\theta(X_{aug},X) -\beta L_{SRAM}(\theta;X,X_{aug})\}.\end{equation}
Therefore, the maximum perturbation loss function is expressed as follows:
\begin{equation}L_{max}=L_{Self-sup}(\theta;X) - L_{con}(\theta;X_{aug},X) -\beta L_{SRAM}(\theta;X_{aug},X).\end{equation}
And in the model loss minimization stage, we expect the model to be better trained on the augmented data by minimizing the self-supervised training loss:
\begin{equation}L_{min}=L_{Self-sup}(\theta;X).\end{equation}

\section{Experiments and Results}
\subsection{Dataset and Implementation Details}
\subsubsection{Dataset:}
We evaluated the model’s performance using the PolypGen dataset \cite{ali2023multi},
which contains colonoscopy data from multiple patients with diverse populations at 6 different medical centers, with a total of 3762 annotated polyp labels. During the training process, we allocate the unlabeled data from five centers ($C2$-$C6$)to the client, representing the process in which each client uses local data for
self-supervised pre-training in federated learning. And we allocate the labeled
data from one of the data centers ($C1$) to the server, indicating the fine-tuning process after pre-training. Specifically, all the local-side data of each client is used as the pre-training dataset. At the same time, we also verified the model performance on an unknown domain CVC-ClinicDB \cite{bernal2015wm}.
\subsubsection{Metrics and Implementation Details:}
We evaluate our model using four segmentation metrics: Mean IoU, Mean Acc, Overall Acc and FreqW Acc used in \cite{Long_2015_CVPR}, among which Mean IoU is the most commonly used segmentation metric. Besides, We choose FedAvg as our basic learning framework, adopt Adam optimization for 300 rounds, set the batch size to 16, and set the learning rate to 3e-4. $\beta$ is set to 2.0 and is discussed in the ablation analysis.

\subsection{Comparison with Prior Work}
We first compare the performance of our method with random initialization and four federated self-supervised learning methods. As shown in Table 1, our method achieves the highest average performance, and although all self-supervised learning methods improve the robustness to data heterogeneity compared with random initialization, the performance of the image mask reconstruction method is higher than that of the contrastive learning method. The MeanIoU value of our method is also 5.91\% and 5.59\% higher than FedBYOL and FedMoco, respectively.  At the same time, it can be seen from experiments that DropPos has the best performance among the three methods based on Vision Transformer. It can be seen from Table \ref{ tab3} that our method exceeds 3.80\% in Mean IoU accuracy compared to using DropPos pre-training alone without a generalization. The results on the unknown domain are similarly shown in Table \ref{ tab2}.

\begin{table}[t]
\centering
\belowrulesep=0pt
\aboverulesep=0pt
\caption{Comparison results on PolypGen $C_1$. The best scores are in bold.}\label{ tab3}
\begin{tabular}{c|cccc}
\toprule
Method&  Mean IoU & Mean Acc &Overall Acc&FreqW Acc \\ \midrule
 Rand. init.(backbone: ResNet50) &0.5591& 0.6585& 0.8823& 0.8143\\
  {FedBYOL}\cite{grill2020bootstrap}&0.5692& 0.6569& 0.8906& 0.8221\\  
 {FedMoco}\cite{he2020momentum} &0.5724& 0.6511& 0.8956& 0.8260\\ 
 \hline
 \hline
  Rand. init.(backbone: ViT)&0.5745& 0.7075& 0.8741& 0.8098\\ 
 {FedMAE}\cite{yan2023label}&0.5891& 0.6989& 0.8877& 0.8222\\
 {FedBEiT}\cite{yan2023label}&0.5882& 0.7013& 0.8861& 0.8207\\  
 {FedDropPos}\cite{wang2024droppos}&0.5903& 0.7119& 0.8826& 0.8172\\ \midrule
Ours&\textbf{0.6283}& \textbf{0.7471}& \textbf{0.9003}& \textbf{0.8400}\\ \bottomrule 
\end{tabular}
\end{table}

\begin{table}[t]
\centering
\belowrulesep=0pt
\aboverulesep=0pt
\caption{Comparison results on an unseen test domain. The best scores are in bold.}\label{ tab2}
\begin{tabular}{c|cccc}
\toprule
Method&  Mean IoU & Mean Acc &Overall Acc&FreqW Acc \\ \midrule
 Rand. init.(backbone: ResNet50)&0.5125& 0.5844& 0.8851& 0.8192\\
  {FedBYOL}\cite{grill2020bootstrap}&0.5297& 0.6129& 0.8866& 0.8234\\   
 {FedMoco}\cite{he2020momentum} &0.5233& 0.6021& 0.8859& 0.8216\\ 
 \hline
 \hline
  Rand. init.(backbone: ViT)&0.5295& 0.5903& 0.9027& 0.8372\\ 
 {FedMAE}\cite{yan2023label}&0.5322& 0.6161& 0.8874& 0.8242\\
 {FedBEiT}\cite{yan2023label}&0.5339& 0.6057& 0.8906& 0.8278\\  
 {FedDropPos}\cite{wang2024droppos}&0.5378& 0.6092& 0.9100& 0.8469\\ \midrule
Ours&\textbf{0.5587}& \textbf{0.6266}& \textbf{0.9084}& \textbf{0.8467}\\ \bottomrule 
\end{tabular}
\end{table}

\begin{figure}[t]
\centering
\includegraphics[width=0.9\textwidth]{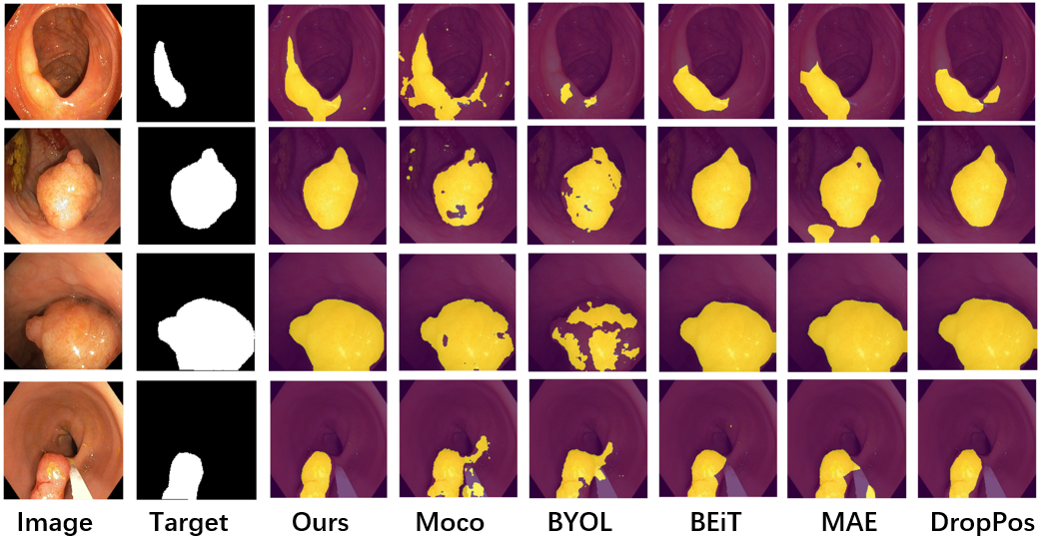}
\caption{Polyp segmentation visualization on the PolypGen dataset.}\label{fig2}
\end{figure}
Figure~\ref{fig2} shows a visualization of the training results of our model and other comparison methods. It can be seen that our method has the most accurate segmentation results.

\subsection{Ablation Studies}
This section conducts ablation experiments to explore the improvement of model performance by SSADA and SRAM.

Table \ref{tab4} shows the effect of SSADA and SRAM in our method. Compared with using only SRAM and using neither SSADA nor SRAM, our method can improve the average IoU accuracy of the model by 1.30\% and 3.80\% respectively. Using only SSADA results in a 2.50\% higher performance compared to using neither SSADA nor SRAM. This indicates that excessive distortion in the enhanced images is successfully alleviated, thereby improving the model's generalization performance. At the same time, for SRAM’s parameter beta, the model’s Mean IoU accuracy reaches its best at 2.0.

\begin{table}[t]
\centering
\belowrulesep=0pt
\aboverulesep=0pt
\caption{Ablation results of SSADA and SRAM. The results of our mode using different $\beta$ are also shown here.}\label{tab4}
\begin{tabular}{c|ccccc}
\toprule
Method&Mean IoU& Mean Acc& Overall Acc&FreqW Acc \\ \midrule
 Rand. init.(backbone: VIT)&0.5745& 0.7075&0.8741& 0.8098\\ 
 Ours w/o SSADA &0.5903& 0.7119& 0.8826& 0.8172\\ 
 Ours w/o SRAM &0.6153& 0.7364& 0.8954& 0.8341\\ \midrule 
Ours(Full model), $\beta = 0.1$&0.6157& 0.7169& 0.9016& 0.8392\\  
 Ours(Full model), $\beta = 0.5$&0.6223& 0.7293& 0.9020& \textbf{0.8406}\\  
 Ours(Full model), $\beta = 1.0$&0.6148& 0.7315& 0.8958& 0.8337\\
 Ours(Full model), $\beta = 2.0$&\textbf{0.6283}& \textbf{0.7471}& 0.9003& 0.8400\\
 Ours(Full model), $\beta = 3.0$&0.6188& 0.7212& \textbf{0.9023}& 0.8403\\
 Ours(Full model), $\beta = 4.0$&0.6156& 0.7228& 0.8995& 0.8373\\
 Ours(Full model), $\beta = 5.0$&0.6083& 0.7156& 0.8967& 0.8336\\
 \bottomrule
\end{tabular}
\end{table} 

%\begin{figure}
%    \begin{minipage}[t]{0.5\linewidth}
%        \centering
%        \includegraphics[width=\textwidth]{result_1.png}
%        \centerline{(a) Comparison with Prior Work}
%    \end{minipage}%
%    \begin{minipage}[t]{0.5\linewidth}
%        \centering
%        \includegraphics[width=\textwidth]{picture/beta.png}
%        \centerline{(b) beta}
%    \end{minipage}
%    \caption{Comparison and Parameter Adjustment}
%\end{figure}

% \subsection{Experiment visualization}

\section{Conclusion}
In this paper, we propose a self-supervised federal learning framework for label-efficient polyp segmentation. We integrate DropPos into a federated learning framework to alleviate label dependence and preserve privacy while using adversarial data augmentation to further enhance generalization capabilities and introduce regularization terms to improve model performance. Experiments show that our method is better than existing federated self-supervised learning methods and effectively improves the generalization ability of the model. 
\subsection*{Acknowledgement}
This work was supported by National Natural Science Foundation of China (Grant No. 62371409)
\subsection*{Disclose}
The authors have no competing interests to declare that are relevant to the comtent of this article.

\newpage
\bibliographystyle{splncs04}
\bibliography{mybibliography.bib}

\begin{thebibliography}{10}
\providecommand{\url}[1]{\texttt{#1}}
\providecommand{\urlprefix}{URL }
\providecommand{\doi}[1]{https://doi.org/#1}

\bibitem{ali2023multi}
Ali, S., Jha, D., Ghatwary, N., Realdon, S., Cannizzaro, R., Salem, O.E., Lamarque, D., Daul, C., Riegler, M.A., Anonsen, K.V., et~al.: A multi-centre polyp detection and segmentation dataset for generalisability assessment. Scientific Data  \textbf{10}(1), ~75 (2023)

\bibitem{bao2021beit}
Bao, H., Dong, L., Piao, S., Wei, F.: Beit: Bert pre-training of image transformers. arXiv preprint arXiv:2106.08254  (2021)

\bibitem{bernal2015wm}
Bernal, J., S{\'a}nchez, F.J., Fern{\'a}ndez-Esparrach, G., Gil, D., Rodr{\'\i}guez, C., Vilari{\~n}o, F.: Wm-dova maps for accurate polyp highlighting in colonoscopy: Validation vs. saliency maps from physicians. Computerized medical imaging and graphics  \textbf{43},  99--111 (2015)

\bibitem{chen2022enhancing}
Chen, C., Qin, C., Ouyang, C., Li, Z., Wang, S., Qiu, H., Chen, L., Tarroni, G., Bai, W., Rueckert, D.: Enhancing mr image segmentation with realistic adversarial data augmentation. Medical Image Analysis  \textbf{82},  102597 (2022)

\bibitem{chen2023meta}
Chen, J., Gao, Z., Wu, X., Luo, J.: Meta-causal learning for single domain generalization. In: Proceedings of the IEEE/CVF Conference on Computer Vision and Pattern Recognition. pp. 7683--7692 (2023)

\bibitem{chen2020simple}
Chen, T., Kornblith, S., Norouzi, M., Hinton, G.: A simple framework for contrastive learning of visual representations. In: International conference on machine learning. pp. 1597--1607. PMLR (2020)

\bibitem{dayan2021federated}
Dayan, I., Roth, H.R., Zhong, A., Harouni, A., Gentili, A., Abidin, A.Z., Liu, A., Costa, A.B., Wood, B.J., Tsai, C.S., et~al.: Federated learning for predicting clinical outcomes in patients with covid-19. Nature medicine  \textbf{27}(10),  1735--1743 (2021)

\bibitem{doersch2015unsupervised}
Doersch, C., Gupta, A., Efros, A.A.: Unsupervised visual representation learning by context prediction. In: Proceedings of the IEEE international conference on computer vision. pp. 1422--1430 (2015)

\bibitem{dou2019domain}
Dou, Q., Coelho~de Castro, D., Kamnitsas, K., Glocker, B.: Domain generalization via model-agnostic learning of semantic features. Advances in neural information processing systems  \textbf{32} (2019)

\bibitem{grill2020bootstrap}
Grill, J.B., Strub, F., Altch{\'e}, F., Tallec, C., Richemond, P., Buchatskaya, E., Doersch, C., Avila~Pires, B., Guo, Z., Gheshlaghi~Azar, M., et~al.: Bootstrap your own latent-a new approach to self-supervised learning. Advances in neural information processing systems  \textbf{33},  21271--21284 (2020)

\bibitem{he2022masked}
He, K., Chen, X., Xie, S., Li, Y., Doll{\'a}r, P., Girshick, R.: Masked autoencoders are scalable vision learners. In: Proceedings of the IEEE/CVF conference on computer vision and pattern recognition. pp. 16000--16009 (2022)

\bibitem{he2020momentum}
He, K., Fan, H., Wu, Y., Xie, S., Girshick, R.: Momentum contrast for unsupervised visual representation learning. In: Proceedings of the IEEE/CVF conference on computer vision and pattern recognition. pp. 9729--9738 (2020)

\bibitem{krull2019noise2void}
Krull, A., Buchholz, T.O., Jug, F.: Noise2void-learning denoising from single noisy images. In: Proceedings of the IEEE/CVF conference on computer vision and pattern recognition. pp. 2129--2137 (2019)

\bibitem{li2018learning}
Li, D., Yang, Y., Song, Y.Z., Hospedales, T.: Learning to generalize: Meta-learning for domain generalization. In: Proceedings of the AAAI conference on artificial intelligence. vol.~32 (2018)

\bibitem{Long_2015_CVPR}
Long, J., Shelhamer, E., Darrell, T.: Fully convolutional networks for semantic segmentation. In: Proceedings of the IEEE Conference on Computer Vision and Pattern Recognition (CVPR) (June 2015)

\bibitem{pathak2016context}
Pathak, D., Krahenbuhl, P., Donahue, J., Darrell, T., Efros, A.A.: Context encoders: Feature learning by inpainting. In: Proceedings of the IEEE conference on computer vision and pattern recognition. pp. 2536--2544 (2016)

\bibitem{pfitzner2021federated}
Pfitzner, B., Steckhan, N., Arnrich, B.: Federated learning in a medical context: a systematic literature review. ACM Transactions on Internet Technology (TOIT)  \textbf{21}(2),  1--31 (2021)

\bibitem{qiao2020learning}
Qiao, F., Zhao, L., Peng, X.: Learning to learn single domain generalization. In: Proceedings of the IEEE/CVF Conference on Computer Vision and Pattern Recognition. pp. 12556--12565 (2020)

\bibitem{qu2022rethinking}
Qu, L., Zhou, Y., Liang, P.P., Xia, Y., Wang, F., Adeli, E., Fei-Fei, L., Rubin, D.: Rethinking architecture design for tackling data heterogeneity in federated learning. In: Proceedings of the IEEE/CVF Conference on Computer Vision and Pattern Recognition. pp. 10061--10071 (2022)

\bibitem{rehman2023federated}
Rehman, M.H.u., Hugo Lopez~Pinaya, W., Nachev, P., Teo, J.T., Ourselin, S., Cardoso, M.J.: Federated learning for medical imaging radiology. The British Journal of Radiology  \textbf{96}(1150),  20220890 (2023)

\bibitem{volpi2018generalizing}
Volpi, R., Namkoong, H., Sener, O., Duchi, J.C., Murino, V., Savarese, S.: Generalizing to unseen domains via adversarial data augmentation. Advances in neural information processing systems  \textbf{31} (2018)

\bibitem{wang2024droppos}
Wang, H., Fan, J., Wang, Y., Song, K., Wang, T., ZHANG, Z.X.: Droppos: Pre-training vision transformers by reconstructing dropped positions. Advances in Neural Information Processing Systems  \textbf{36} (2024)

\bibitem{yan2023label}
Yan, R., Qu, L., Wei, Q., Huang, S.C., Shen, L., Rubin, D., Xing, L., Zhou, Y.: Label-efficient self-supervised federated learning for tackling data heterogeneity in medical imaging. IEEE Transactions on Medical Imaging  (2023)

\bibitem{zhang2016colorful}
Zhang, R., Isola, P., Efros, A.A.: Colorful image colorization. In: Computer Vision--ECCV 2016: 14th European Conference, Amsterdam, The Netherlands, October 11-14, 2016, Proceedings, Part III 14. pp. 649--666. Springer (2016)

\bibitem{zhao2020maximum}
Zhao, L., Liu, T., Peng, X., Metaxas, D.: Maximum-entropy adversarial data augmentation for improved generalization and robustness. Advances in Neural Information Processing Systems  \textbf{33},  14435--14447 (2020)

\end{thebibliography}

%\end{thebibliography}
\end{spacing}
\end{sloppypar}%防止行溢出页边距
\end{document}